%% file: main.tex
\documentclass[conference]{IEEEtran}
\IEEEoverridecommandlockouts
\usepackage{cite}
\usepackage{amsmath,amssymb,amsfonts}
\usepackage{algorithmic}
\usepackage{graphicx}
\usepackage{textcomp}
\usepackage[table]{xcolor}
\usepackage{pifont}
\usepackage{tikz}
\usepackage{hyperref}
\usepackage{subcaption} 
\usepackage{graphicx}   

\usepackage{booktabs}
\def\BibTeX{{\rm B\kern-.05em{\sc i\kern-.025em b}\kern-.08em
    T\kern-.1667em\lower.7ex\hbox{E}\kern-.125emX}}
\begin{document}

\title{Dual-R-DETR: Resolving Query Competition with Pairwise Routing in Transformer Decoders}

\author{\IEEEauthorblockN{Ye Zhang$^{1}$, Qi Chen$^{2}$, Wenyou Huang$^{3}$, Rui Liu$^{4}$, Zhengjian Kang$^{5,*}$}
\IEEEauthorblockA{
$^{1}$University of Pittsburgh, $^{2}$University of California, Irvine, $^{3}$Stevens Institute of Technology\\
$^{4}$Illinois Institute of Technology, $^{5}$New York University\\
$^{1}$yez12@pitt.edu, $^{2}$qic7@uci.edu, $^{3}$nealhuang0.0@gmail.com, $^{4}$liuruiabc1@gmail.com, $^{5}$zk299@nyu.edu}}

\maketitle
\footnotetext[1]{Corresponding author. Email: zk299@nyu.edu}

\begin{abstract}
Detection Transformers (DETR) formulate object detection as a set prediction problem and enable end-to-end training without post-processing. However, object queries in DETR interact through symmetric self-attention, which enforces uniform competition among all query pairs. This often leads to inefficient query dynamics, where multiple queries converge to the same object while others fail to explore alternative regions.
We propose \textbf{Dual-R-DETR}, a competition-aware DETR framework that explicitly regulates query interactions via \emph{pairwise routing} in transformer decoders. Dual-R-DETR distinguishes query-to-query relations as either competitive or cooperative based on appearance similarity, prediction confidence, and spatial geometry. It introduces two complementary routing behaviors: \emph{suppressor routing} to attenuate interactions among queries targeting the same object, and \emph{delegator routing} to encourage diversification across distinct regions. These behaviors are realized through lightweight, learnable low-rank biases injected into decoder self-attention, enabling asymmetric query interactions while preserving the standard attention formulation.
To ensure inference efficiency, routing biases are applied only during training using a dual-branch strategy, and inference reverts to vanilla self-attention with no additional computational cost. Extensive experiments on COCO and Cityscapes demonstrate that Dual-R-DETR consistently improves multiple DETR variants, outperforming DINO by 1.7\% mAP with a ResNet-50 backbone and achieving 57.6\% mAP with Swin-L under comparable settings. 
Code is available at \url{https://github.com/YZk67/Dual-R-DETR}.
\end{abstract}



\input{sec/01_introduction}
\input{sec/02_related_work}

\input{sec/03_method}

\input{sec/04_experiments}

\input{sec/05_conclusion}

\bibliographystyle{IEEEbib}
\bibliography{icme2026references}


\end{document}

%% file: sec/01_introduction.tex
\section{Introduction}
Object detection~\cite{wang2025uniocc} is a fundamental task in computer vision, aiming to recognize and localize objects in images. Detection Transformers (DETR)~\cite{detr} reformulate object detection as a set prediction problem and eliminate hand-crafted components such as non-maximum suppression (NMS), enabling end-to-end optimization with learnable object queries.

\begin{figure}[!t]
\begin{minipage}[b]{1.0\linewidth}
  \centering
  \centerline{\includegraphics[width=8.5cm,height=2.6cm]{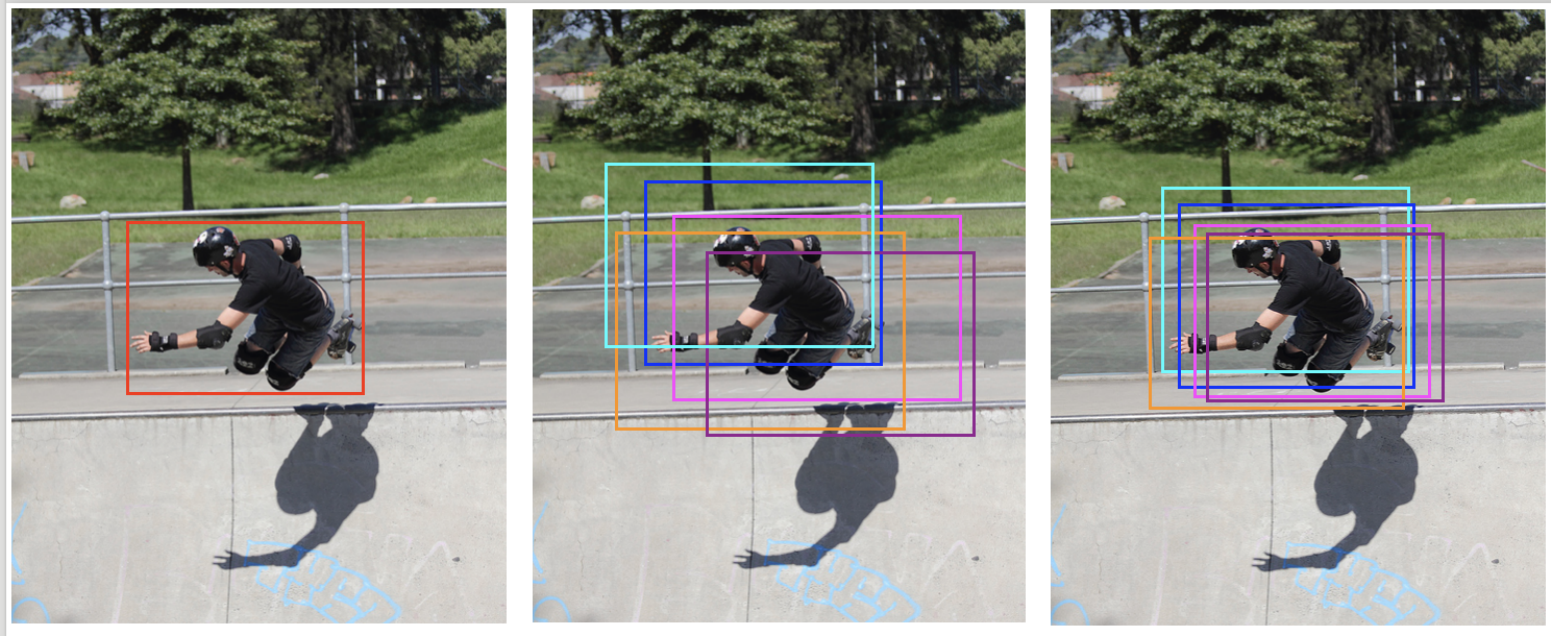}}
  \vspace{-0.1cm}
  \begin{tabular*}{8.5cm}{@{}p{2.83cm}@{}p{2.83cm}@{}p{2.83cm}@{}}
  \centering (a) & \centering (b) & \centering (c) \\
  \end{tabular*}
\end{minipage}
    \caption{Query competition evolution in DETR decoder layers. (a) Ground truth objects; (b) initial query predictions before decoder processing; (c) query predictions after decoder layers, where multiple queries converge to the same targets.}
    \vspace{-0.6cm}
    \label{fig:res}
\end{figure}

A defining characteristic of DETR is its ability to produce non-duplicate detections through one-to-one label assignment~\cite{detr,deformable,gao2024ease}. However, this exclusivity is enforced only at the output level. During decoding, object queries interact through self-attention and cross-attention, leading to implicit competition throughout feature refinement. As illustrated in Fig.~\ref{fig:res}, multiple queries are often attracted to the same target objects during early decoder layers, while other queries fail to explore alternative regions. Although redundant queries are eventually suppressed by Hungarian matching, this competition is resolved implicitly and late, resulting in inefficient query utilization and unstable optimization.

This behavior reveals a fundamental limitation of standard DETR decoders: query-to-query interactions are modeled by symmetric self-attention, which applies the same interaction mechanism to all query pairs. Such uniform treatment cannot simultaneously suppress redundant queries converging to the same object and encourage complementary queries to coordinate over distinct regions. In particular, symmetric attention induces mutual reinforcement between competing queries, causing them to repeatedly focus on the same targets and delaying effective competition resolution.

Recent efforts have explored improving DETR from various perspectives, including deformable attention~\cite{deformable}, denoising training~\cite{li2022dn}, and enhanced query representations~\cite{dino,hu2023dac,kang2025lp,kang2026paq}. While these approaches improve convergence speed or representation quality, they largely retain the standard decoder self-attention mechanism and do not explicitly regulate inter-query interactions during feature refinement.

Several recent works have also investigated query dynamics more directly by modifying query interactions or selection strategies. For example, EASE-DETR~\cite{gao2024ease} introduces early suppression mechanisms to mitigate redundant query responses, and query selection-based approaches~\cite{senthivel2024qr} aim to improve efficiency by pruning less informative queries during decoding. These methods operate at the level of query selection or lifecycle management. In contrast, our approach does not remove or terminate queries. Instead, Dual-R-DETR explicitly regulates \emph{pairwise query--query interactions} throughout decoding, selectively suppressing competitive interactions while preserving cooperative interactions among complementary queries.

Motivated by this observation, we propose \textbf{Dual-R-DETR}, which explicitly resolves query competition by introducing adaptive \emph{pairwise routing} into decoder self-attention. Instead of treating all query pairs uniformly, Dual-R-DETR learns to modulate query interactions based on readily available signals, including appearance similarity, prediction confidence, and spatial geometry. It introduces two complementary routing behaviors: suppressor routing, which attenuates interactions between competing queries targeting the same object, and delegator routing, which promotes diversification by encouraging queries to explore different regions. Prior works manage which queries live or die, but we model how surviving queries interact via asymmetric pairwise modulation.

Our main contributions are summarized as follows:
\begin{itemize}
\item \textbf{Competition-Aware Query Routing}: We introduce a pairwise routing mechanism that explicitly distinguishes competitive and cooperative query interactions in DETR decoders.
\item \textbf{Dual Routing Behaviors}: We propose suppressor and delegator routing to respectively attenuate redundant query interactions and encourage exploration of complementary regions.
\item \textbf{Efficient Training Strategy}: We employ dual-branch training to incorporate routing during training while incurring no additional inference cost.
\end{itemize}

Extensive experiments on COCO~\cite{coco} and Cityscapes~\cite{Cordts2016} for object detection and instance segmentation demonstrate that Dual-R-DETR consistently improves multiple DETR baselines across different backbones.

%% file: sec/02_related_work.tex
\section{Related Work}

\subsection{DETR Variants and Query Competition.}
Detection Transformer (DETR)~\cite{detr} casts object detection as a set prediction problem with a fixed number of learnable queries and one-to-one Hungarian matching, enabling end-to-end training without non-maximum suppression. Subsequent works mainly improve convergence, stability, and representation quality. Deformable DETR~\cite{deformable} introduces multi-scale deformable attention and encoder-driven proposals, while DN-DETR~\cite{li2022dn} and DINO~\cite{dino} stabilize optimization through denoising and contrastive objectives. DAB-DETR~\cite{liu2022dab} injects spatial priors via anchor parameterization.
Despite these advances, a fundamental issue persists: \emph{inefficient query competition}. During decoding, cross-attention may pull multiple queries toward the same object, yet one-to-one matching provides positive supervision to only a single query, leaving other well-localized queries without positive signals. This mismatch leads to redundant computation and underutilized query capacity, while standard decoder self-attention does not explicitly distinguish competitive versus cooperative query interactions.

\subsection{Query Routing and Interaction Modeling.}
Several recent methods investigate query dynamics by modifying query interactions or selection strategies. EASE-DETR~\cite{gao2024ease} introduces query-level routing and suppression mechanisms during decoding, while QR-DETR~\cite{senthivel2024qr} explores query reassignment and early termination to reduce ineffective query updates. These approaches primarily operate at the level of \emph{query selection or lifecycle control}, deciding which queries should be retained or suppressed, rather than explicitly modeling how different queries \emph{interact} with each other.
From a different perspective, DAC-DETR~\cite{hu2023dac} decouples attention layers using parallel decoders to improve training efficiency, but still treats self-attention as a uniform operation without distinguishing diverse query-to-query relationships.

In contrast, \textbf{Dual-R-DETR} focuses on \emph{pairwise query interactions} within decoder self-attention. Instead of selecting or pruning queries, we differentiate competing query pairs from complementary ones and modulate their interactions through asymmetric suppressor and delegator routes. These routing biases are introduced only during training, while inference preserves standard self-attention without additional computational overhead.
\vspace{-0.3cm}

\begin{figure*}[!t]
\centering
\includegraphics[width=1.0\textwidth,height=0.35\textheight]{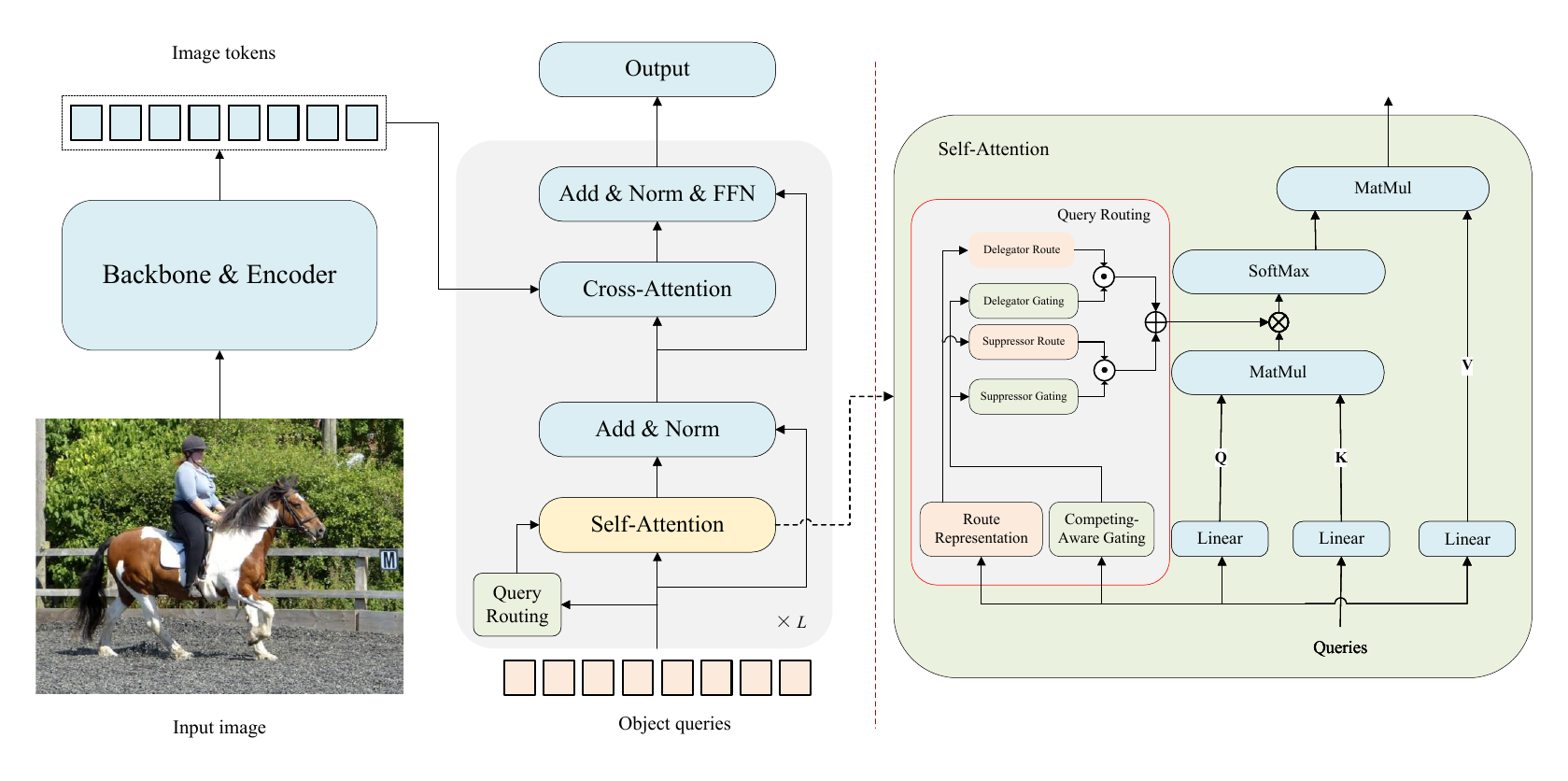}
\caption{Dual-R-DETR framework overview. It incorporates dual routing mechanisms---suppressor and delegator routes---into decoder self-attention layers to address query competition through learnable pairwise biases.}
\vspace{-0.5cm}
\label{fig:overview}
\end{figure*}

\subsection{Assignment Strategies and Structured Pairwise Biases.}
Another line of work revisits label assignment to increase supervision density. Group-DETR~\cite{chen2023group}, H-DETR~\cite{hdetr}, and Stable-DETR~\cite{liu2023stable} introduce one-to-many or hybrid matching strategies, while Co-DETR~\cite{zong2023codetr} combines one-to-one and one-to-many branches through collaborative training. These methods mainly modify \emph{how many} queries are supervised per object, but often leave the underlying query interaction mechanism unchanged.
Our approach is orthogonal to assignment design and can be combined with such strategies. Dual-R-DETR reshapes query dynamics through \emph{structured pairwise attention biases}. Inspired by low-rank representations~\cite{wang2021a,wang2004face}, we parameterize pairwise routing biases with low-rank factorization, enabling fine-grained, competition-aware interaction modeling during training while preserving standard self-attention at inference with no additional computational cost.
\vspace{-0.5cm}

%% file: sec/03_method.tex
\section{Method}
\label{sec:methodology}

\subsection{DETR Architecture}
DETR-style detectors consist of three main components: a backbone network, a transformer encoder--decoder, and prediction heads. The backbone extracts tokens $\mathbf{X} \in \mathbb{R}^{m \times d}$ from an input image $\mathbf{I} \in \mathbb{R}^{h \times w \times 3}$, where $m$ denotes the number of tokens and $d$ the embedding dimension. The encoder enhances contextual representations via self-attention as
\begin{equation}
\mathbf{O} = \text{Encoder}(\mathbf{X}) \in \mathbb{R}^{m \times d}.
\end{equation}

The decoder processes a fixed set of $n$ learnable object queries $\mathbf{Q} = \{q_1, \ldots, q_n\} \in \mathbb{R}^{n \times d}$ through $L$ sequential layers. Each layer consists of self-attention (query--query interactions), cross-attention (query--image interactions), and a feed-forward network (FFN). Notably, decoder self-attention models query--query interactions symmetrically, implicitly assuming that all query pairs benefit equally from mutual interaction.

After $L$ decoding layers, the refined queries $\mathbf{Q}^L$ are fed into prediction heads to produce bounding boxes and classification scores:
\begin{equation}
\mathbf{Q}^l = \text{FFN}\big(\text{CrossAttn}(\text{SelfAttn}(\mathbf{Q}^{l-1}), \mathbf{O})\big),
\label{eq:inference1}
\end{equation}
\begin{equation}
\mathbf{Y} = \text{Head}(\mathbf{Q}^{L}),
\label{eq:inference2}
\end{equation}
where $\mathbf{Y} = \{(b_i, c_i)\}_{i=1}^n$ denotes the predicted bounding boxes $b_i \in \mathbb{R}^4$ and class scores $c_i$.

\subsection{Dual-R-DETR Overview}
Object detection with DETR inherently involves competition among queries: multiple queries may converge to similar spatial regions and compete to represent the same object. In such cases, some query interactions should be suppressive to avoid duplicate detections, while others should be facilitative to encourage coverage of complementary regions. However, standard self-attention treats all query pairs symmetrically, a uniform interaction assumption that fundamentally conflicts with these opposing requirements. For example, if one query suppresses another to reduce redundancy, symmetric attention would enforce the reverse interaction as well, leading to mutual suppression that hinders effective query specialization.

Dual-R-DETR addresses this structural mismatch by introducing competition-aware routing into decoder self-attention. Specifically, we inject learnable pairwise routing biases into the pre-softmax attention logits to modulate query--query interactions. As illustrated in Figure~\ref{fig:overview}, we preserve the standard encoder--decoder architecture while augmenting self-attention with adaptive routing that differentiates between competing and complementary query pairs.

To preserve inference efficiency and stability, routing is applied only during training via a dual-branch strategy. Intuitively, routing biases guide query specialization during training, while the learned representations are distilled into a standard branch, enabling stable inference without routing overhead.

\subsection{Query Routing Mechanism}

\noindent\textbf{Pairwise Routing Formulation.}
Query competition arises from interactions between specific query pairs rather than from individual queries in isolation. When multiple queries attend to the same object, their mutual interactions should be inhibited, whereas queries focusing on distinct regions should be encouraged to coordinate. Consequently, routing decisions must be defined at the level of pairwise query interactions rather than applied uniformly to individual queries.

\noindent\textbf{Low-Rank Pairwise Route Representation.}
Query competition involves two fundamentally different interaction requirements: suppressing redundant queries that converge to the same object, and reinforcing complementary queries that explore distinct regions. These opposing effects cannot be captured by a single routing signal. We therefore explicitly model two routing types—\emph{suppressor} and \emph{delegator}—to separately modulate inhibitory and facilitative interactions.

To enable adaptive routing, we construct a routing representation for each query at each decoder layer:
\begin{equation}
z_i = \phi([q_i \| pos_i]) \in \mathbb{R}^{d_z},
\end{equation}
where $q_i$ denotes the query feature, $pos_i$ the positional encoding, and $\phi$ a learnable projection. Let $\mathbf{Z} = [z_1, \ldots, z_n]^\top \in \mathbb{R}^{n \times d_z}$. We encode the two routing types using low-rank factorization:
\begin{equation}
\Delta^{\text{sup}} = U^{\text{sup}}(\mathbf{Z}) V^{\text{sup}}(\mathbf{Z})^\top, \quad
\Delta^{\text{del}} = U^{\text{del}}(\mathbf{Z}) V^{\text{del}}(\mathbf{Z})^\top,
\end{equation}
where $U^{*}(\mathbf{Z}) = \mathbf{Z}\mathbf{W}_U^{*} \in \mathbb{R}^{n \times r}$ and $V^{*}(\mathbf{Z}) = \mathbf{Z}\mathbf{W}_V^{*} \in \mathbb{R}^{n \times r}$ with low rank $r \ll d_z$.  
Low-rank factorization reduces the routing parameter space from $\mathcal{O}(n^2)$ to $\mathcal{O}(nr)$, making pairwise routing computationally practical for hundreds of queries. It also encourages queries with similar representations to share routing patterns, enabling consistent interaction behaviors across query groups.

\noindent\textbf{Competition-Aware Pairwise Gating.}
To distinguish suppressive versus facilitative interactions, we define query descriptors $\mathbf{x}_i = [s_i, c_i, g_i]$ using interaction-relevant signals: mean cosine similarity $s_i$ to other queries, predicted classification confidence $c_i$, and log-scale geometric area $g_i$. These signals are readily available during decoding and jointly capture appearance redundancy, detection strength, and spatial overlap. Pairwise gating is implemented via a bilinear interaction:
\begin{equation}
p_{\text{sup}}(i,j) = \sigma(\mathbf{a}_i^\top \mathbf{b}_j), \quad
p_{\text{del}}(i,j) = 1 - p_{\text{sup}}(i,j),
\end{equation}
where $\mathbf{a}_i = \mathbf{x}_i \mathbf{W}_a$ and $\mathbf{b}_j = \mathbf{x}_j \mathbf{W}_b \in \mathbb{R}^{r_g}$.  
The bilinear formulation enables asymmetric routing decisions between query pairs, for example allowing one query to suppress another without enforcing the reverse interaction, which simple similarity measures or shared MLPs cannot express. Enforcing $p_{\text{del}} = 1 - p_{\text{sup}}$ ensures mutually exclusive routing decisions, preventing conflicting modulation signals for the same query pair and simplifying optimization.

\noindent\textbf{Attention Bias Integration.}
We introduce asymmetric, signed attention biases to explicitly break the symmetry of standard self-attention:
\begin{equation}
\gamma_{\text{sup}} = -\text{softplus}(\tilde{\gamma}_{\text{sup}}), \quad
\gamma_{\text{del}} = +\text{softplus}(\tilde{\gamma}_{\text{del}}).
\end{equation}
The sign constraint is not pre-assigned based on semantic roles; instead, it enforces opposite interaction effects that are learned end-to-end from data. The routed bias is computed as
\begin{equation}
\mathbf{B} = p_{\text{sup}} \odot (\gamma_{\text{sup}} \Delta^{\text{sup}}) + p_{\text{del}} \odot (\gamma_{\text{del}} \Delta^{\text{del}}),
\end{equation}
where $\odot$ denotes element-wise multiplication. The routing bias acts as a soft modulation rather than a hard constraint, allowing robustness to imperfect gating decisions.

The modified self-attention is then given by
\begin{equation}
\text{SelfAttn} = \text{Softmax}\left(\frac{\mathbf{Q}\mathbf{K}^\top}{\sqrt{d}} + \mathbf{B}\right),
\label{eq:selfattn}
\end{equation}
where injecting routing biases into pre-softmax logits preserves the probabilistic structure of attention while enabling smooth and stable modulation of interaction strengths. The bias $\mathbf{B} \in \mathbb{R}^{n \times n}$ is computed per layer and per sample, and broadcast across batch and head dimensions.

\noindent\textbf{Routing Summary.}
In summary, the proposed routing mechanism resolves query competition through three complementary components: a low-rank formulation that enables efficient pairwise modeling, competition-aware gating that identifies suppressive versus facilitative interactions, and asymmetric bias injection that selectively modulates query--query interactions. Together, these components reduce redundant query refinement while preserving beneficial interactions, leading to more effective utilization of the fixed set of queries.

\subsection{Dual-Branch Training Strategy}
Introducing competition-aware routing into decoder self-attention enables effective query specialization, but may introduce additional computational cost if applied at test time. To ensure efficient and stable inference behavior, we adopt a dual-branch training strategy that decouples training-time routing from inference-time execution.

Both branches share the same backbone, encoder, decoder, and prediction heads, and optimize the same detection objective. The two branches differ only in decoder self-attention: the main branch employs standard attention and serves as the inference-time surrogate, while the auxiliary branch incorporates routing-augmented attention as defined in Eq.~\ref{eq:selfattn}. The auxiliary branch introduces no additional supervision; it solely modulates interaction dynamics to encourage specialization. While prior DETR-based works~\cite{hu2023dac} adopt dual branches for different purposes, we employ this strategy specifically to isolate competition-aware routing during training while preserving standard inference behavior.

The overall loss is defined as
\begin{equation}
\mathcal{L} = \mathcal{L}_{\text{main}} + \alpha_t \cdot \mathcal{L}_{\text{aux}},
\label{eq:loss}
\end{equation}
where $\alpha_t$ is gradually increased from $\alpha_{\min}$ to $\alpha_{\max}$ using a cosine warm-up schedule. During inference, only the main branch is retained, resulting in zero additional computational overhead compared to standard DETR models.

%% file: sec/04_experiments.tex
\section{Experiments}
\label{sec:experiments}

\begin{table*}[!t] 
  \caption{The performance on COCO \texttt{val}2017 over different baselines using various backbones, epochs, and queries. Rows marked with (ours) show results with our routing module.}
  \setlength{\tabcolsep}{4pt} 
  \small
  \centering
  \begin{tabular}{l|c|cc|cccccc}
  \hline
    Method & Backbone & Epochs & Queries & mAP & AP$_{50}$ & AP$_{75}$ & AP$_S$ & AP$_M$ & AP$_L$ \\ 
  \hline\hline
    Deformable-DETR++~\cite{deformable} & RN-50 & 12 & 300 & 46.8 & 65.6 & 51.1 & 30.1 & 50.4 & 60.3 \\
    Deformable-DETR++~\cite{deformable} & RN-50 & 36 & 300 & 49.0 & 67.6 & 53.5 & 32.6 & 52.3 & 63.3 \\
    Deformable-DETR++~\cite{deformable} & Swin-T & 12 & 300 & 49.3 & 67.9 & 53.6 & 31.6 & 52.4 & 64.3 \\
    Deformable-DETR++ (ours) & RN-50 & 12 & 300 & 48.1({+1.3}) & 66.4({+0.8}) & 52.5({+1.4}) & 31.6({+1.5}) & 51.8({+1.4}) & 62.5({+2.2}) \\
    Deformable-DETR++ (ours) & RN-50 & 24 & 300 & 49.4({+0.4}) & 67.9({+0.3}) & 53.9({+0.4}) & 33.1({+0.5}) & 52.6({+0.3}) & 64.0({+0.7}) \\
    Deformable-DETR++ (ours) & Swin-T & 12 & 300 & 50.1({+0.8}) & 69.0({+1.1}) & 54.4({+0.8}) & 32.9({+1.3}) & 53.3({+0.9}) & 64.8({+0.5}) \\
  \hline
    DAB-Def-DETR++~\cite{liu2022dab} & RN-50 & 12 & 300 & 48.0 & 66.2 & 52.4 & 31.9 & 51.4 & 61.7 \\
    DAB-Def-DETR++ (ours) & RN-50 & 12 & 300 & 48.7({+0.7}) & 67.0({+0.8}) & 52.9({+0.5}) & 33.1({+1.2}) & 52.0({+0.6}) & 62.8({+1.1}) \\
    \hline
    DINO~\cite{dino} & RN-50 & 12 & 900 & 49.0 & 66.7 & 53.5 & 32.1 & 52.7 & 63.2 \\
    DINO (ours) & RN-50 & 12 & 900 & 50.1({+1.1}) & 67.9({+1.2}) & 54.8({+1.3}) & 33.9({+1.8}) & 53.7({+1.0}) & 63.8({+0.6}) \\
  \hline
  \end{tabular}
  \label{tab:coco2017-comparision}
\end{table*}

\begin{table*} 
  \caption{The performance on COCO \texttt{val}2017 with SOTA methods using different backbones.}
  \setlength{\tabcolsep}{11pt} 
  \small
  \centering
  \begin{tabular}{l|c|cc|cccccc}
  \hline
    Method & Backbone & Epochs & Queries & mAP & AP$_{50}$ & AP$_{75}$ & AP$_S$ & AP$_M$ & AP$_L$ \\ 
  \hline\hline
    DN-Def-DETR++~\cite{li2022dn} & ResNet-50 & 12 & 900 & 48.7 & 66.4 & 52.9 & 32.1 & 52.1 & 63.7 \\
    H-DETR~\cite{hdetr} & ResNet-50 & 12 & 900  & 48.7 & 66.4 & 52.9 & 31.2 & 51.5 & 63.5 \\
    Group-DETR~\cite{chen2023group} & ResNet-50 & 12 & 900  & 49.8 & - & - & 32.4 & 53.0 & 64.2 \\
    DAC-DETR~\cite{hu2023dac} & ResNet-50 & 12 & 900  & 50.0 & 67.6 & 54.7 & 32.9 & 53.1 & 64.4 \\
    Salience-DETR~\cite{hou2024salience} & ResNet-50 & 12 & 900  & 49.2 & 67.1 & 53.8 & 32.7 & 53.0 & 63.1 \\
    Rank-DETR~\cite{pu2023rank} & ResNet-50 & 12 & 900  & 50.4 & 67.9 & 55.2 & 33.6 & 53.8 & 64.2 \\
    DINO~\cite{dino} & ResNet-50 & 12 & 900  & 49.0 & 66.7 & 53.5 & 32.1 & 52.7 & 63.2 \\
    Dual-R-DETR & ResNet-50 & 12 & 900  & 50.7 & 68.8 & 55.0 & 34.1 & 54.2 & 64.6 \\
  \hline
    H-DETR~\cite{hdetr} &  Swin-L & 12 & 900  & 55.9 & 75.2 & 61.0 & 39.1 & 59.9 & 72.2 \\
    DAC-DETR~\cite{hu2023dac} &  Swin-L & 12 & 900  & 57.3 & 75.7 & 62.7 & 40.1 & 61.5 & 74.4 \\
    Salience-DETR~\cite{hou2024salience} &  Swin-L & 12 & 900  & 56.5 & 75.0 & 61.5 & 40.2 & 61.2 & 72.8 \\
    DINO~\cite{dino} &  Swin-L & 12 & 900  & 56.8 & 75.4 & 62.0 & 40.1 & 60.5 & 73.2 \\
    Dual-R-DETR & Swin-L & 12 & 900 & 57.6 & 76.4 & 63.2 & 40.8 & 61.8 & 73.6 \\
  \hline
  \end{tabular}
  \label{tab:coco2017}
  \vspace{-0.3cm}
\end{table*}

\subsection{Implementation details}
We select two backbones to evaluate our approach: ResNet-50~\cite{he2016deep} pretrained on ImageNet-1k and Swin-Large~\cite{liu2021swin} pretrained on ImageNet-22k~\cite{deng2009imagenet}. All models are trained using the AdamW optimizer~\cite{loshchilov2017decoupled} with an initial learning rate of $2e^{-4}$ and weight decay of $1e^{-4}$. We adopt standard training schedules of 1$\times$ (12 epochs) and 2$\times$ (24 epochs), where the learning rate is reduced by a factor of 0.1 at the 11th and 20th epochs, respectively. Training is performed with a batch size of 16 across 8 NVIDIA RTX 3090 GPUs. Each experiment is averaged over 3 random seeds. Following established practices~\cite{detr}, we apply standard data augmentations, including random resizing, cropping, and horizontal flip during training. 
The hyperparameters in low-rank computation in our experiments is configured as follows $d_z=16$, $r=16$, and $r_g=32$. For other parameters, we use default parameter settings in Deformable-DETR++~\cite{deformable}.

\subsection{Datasets and Metrics}
We perform comprehensive experiments to evaluate our model across multiple benchmarks and tasks. We use COCO 2017~\cite{coco} for object detection. We further extend our evaluation to instance segmentation using COCO 2017~\cite{coco} and CityScapes 2016~\cite{Cordts2016}. Following established protocol~\cite{detr}, we evaluate detection performance using standard COCO metrics, including mean Average Precision (mAP) at different IoU thresholds (0.5, 0.75, and 0.5:0.95), as well as different scales across small, medium, and large objects. For instance segmentation, we consider the metrics from both mask mAP and box mAP metrics, respectively.

\subsection{Quantitative Results}
We evaluate Dual-R-DETR across multiple strong baselines on COCO 2017~\cite{coco}, as shown in Table~\ref{tab:coco2017-comparision}. Our query routing module consistently improves performance across different architectures, backbones, training schedules, and query configurations, demonstrating strong generalizability.
For Deformable-DETR++, integrating our routing mechanism into decoder self-attention yields consistent gains. On ResNet-50 with 12 epochs, we observe a +1.3\% mAP improvement (48.1\% vs. 46.8\%), and similar gains are achieved on Swin-T (+0.8\% mAP). Notably, the 24-epoch Dual-R-DETR result (49.4\% mAP) matches or slightly exceeds the baseline trained for 36 epochs (49.0\% mAP), effectively reducing training time while maintaining comparable accuracy. Improvements are also consistent across AP metrics, particularly for medium and large objects.
When applied to other DETR variants, Dual-R-DETR delivers consistent gains across architectures. DAB-Def-DETR++ achieves a +0.7\% mAP improvement, while DINO shows a +1.1\% mAP gain, indicating that the proposed pairwise routing is architecture-agnostic.

To compare with state-of-the-art methods, we integrate our routing module with DINO~\cite{dino} and adopt the one-to-many training strategy from Stable-DETR~\cite{liu2023stable}, as shown in Table~\ref{tab:coco2017}. This combination yields a +1.7\% mAP improvement on ResNet-50 (50.7\% vs. 49.0\%). With a Swin-L backbone, Dual-R-DETR achieves 57.6\% mAP, surpassing recent state-of-the-art methods including DAC-DETR~\cite{hu2023dac} and Salience-DETR~\cite{hou2024salience}.

\subsection{Ablation Study}
Table~\ref{tab:abl:component} demonstrates the individual and combined contributions of our proposed components using the Deformable-DETR++ baseline. The suppressor route alone achieves substantial 1.0\% mAP improvement (47.8\%) with consistent gains across all metrics, while the delegator route contributes a modest 0.4\% mAP improvement (47.2\%). When combined, we observe a synergistic effect with 1.3\% mAP improvement (48.1\%), exceeding the sum of individual contributions.
This indicates complementary operation: the suppressor route reduces competitive interference between similar queries, while the delegator route enhances coverage by encouraging non-competing queries to explore different regions. The learned routing gate dynamically balances these effects, preventing over-suppression while maintaining competition control. This validates our dual-route design, as the combined framework achieves a higher mAP than either component in isolation.

\begingroup
\setlength{\tabcolsep}{3pt} 
\renewcommand{\arraystretch}{1.0} 
\begin{table}[t]
\caption{Ablation study on proposed components using Deformable-DETR++ as baseline with backbone ResNet-50. S: suppressor route, D: delegator router.}
  \centering
  \begin{tabular}{cc|ccc}
  \hline
    S & D & mAP & AP$_{50}$ & AP$_{75}$ \\ 
  \hline\hline
    & & 46.8 & 65.6 & 51.1 \\
    \checkmark & & 47.8({+1.0}) & 66.2({+0.6}) & 52.1({+1.0}) \\
    &  \checkmark & 47.2({+0.4}) & 65.9({+0.3}) & 51.3({+0.2}) \\
    \checkmark &  \checkmark & 48.1({+1.3}) & 66.4({+0.8}) & 52.5({+1.4}) \\
  \hline
\end{tabular}
\label{tab:abl:component}
\end{table}
\endgroup

\subsection{Instance Segmentation}
We extend Dual-R-DETR to instance segmentation by adding a mask prediction head to the transformer. Using Deformable-DETR~\cite{deformable} as the baseline, we evaluate on COCO val2017~\cite{coco} and Cityscapes~\cite{Cordts2016} under different training schedules. As shown in Table~\ref{tab:insseg}, our query routing consistently improves both mask and box mAP across datasets and schedules. On COCO val2017, Dual-R-DETR improves mask mAP by 1.4\% and 1.9\% at 12 and 24 epochs, respectively. Similar gains are observed on Cityscapes, with mask mAP improvements of 1.4\% and 1.6\%. Box mAP also improves consistently across both datasets, indicating that the proposed routing benefits both localization and segmentation.
\vspace{-0.1cm}

\begingroup
\setlength{\tabcolsep}{4pt} 
\renewcommand{\arraystretch}{1.0} 
\begin{table}[t]
  \centering
  \footnotesize
  \caption{Instance segmentation on COCO \texttt{val}2017 and CityScapes 2016 with backbone ResNet-50.}
  \begin{tabular}{l|c|c|c}
  \hline
    Method & Epochs & Mask mAP & Box mAP \\
  \hline\hline
  \multicolumn{4}{l}{\texttt{\textbf{Dataset: COCO val2017}}} \\
  \hline
  Deformable-DETR~\cite{deformable} & 12 & 32.4 & 46.5 \\
  Deformable-DETR~\cite{deformable} & 24 & 35.1 & 48.6 \\
  Dual-R-DETR & 12 & 33.8({+1.4}) & 48.1({+1.6}) \\
  Dual-R-DETR & 24 & 37.0({+1.9}) & 49.8({+1.2}) \\
  \hline\hline
  \multicolumn{4}{l}{\texttt{\textbf{Dataset: CityScapes 2016}}} \\
  \hline
  Deformable-DETR~\cite{deformable} & 12 & 34.8 & 52.6 \\
  Deformable-DETR~\cite{deformable} & 24 & 36.6 & 54.4 \\
  Dual-R-DETR & 12 & 36.2({+1.4}) & 53.8({+1.2}) \\
  Dual-R-DETR & 24 & 38.2({+1.6}) & 55.9({+1.5}) \\
  \hline
  \end{tabular}
\label{tab:insseg}
\vspace{-0.3cm}
\end{table}
\endgroup

\subsection{Inference Efficiency}
As shown in Table~\ref{tab:efficiency}, Dual-R-DETR achieves comparable inference efficiency to existing DETR variants. The number of parameters, FLOPs, and runtime FPS are comparable across methods, indicating that the proposed competition-aware routing introduces no additional inference overhead.

\begin{table}[t]
\centering
\caption{Inference Efficiency Comparison. All methods are evaluated under the same inference settings.}
\begin{tabular}{lccc}
\toprule
Method & Params (M) & FLOPs (G) & FPS $\uparrow$ \\
\midrule
Deformable-DETR~\cite{deformable} & 40.5 & 188 & 16.4 \\
DN-DETR~\cite{li2022dn} & 41.2 & 192 & 16.6 \\
DINO~\cite{dino} & 42.0 & 196 & 15.8 \\
\midrule
Dual-R-DETR (Ours) & 41.3 & 190 & 16.5 \\
\bottomrule
\end{tabular}
\label{tab:efficiency}
\vspace{-0.5cm}
\end{table}

%% file: sec/05_conclusion.tex

\section{Conclusion}
We presented Dual-R-DETR, a method that addresses query competition in DETR-style detectors through suppressor and delegator routing. Our method consistently improves multiple COCO baselines without inference overhead. Future work will extend routing to other transformer architectures and explore adaptive routing schedules.